\documentclass[runningheads]{llncs}

\usepackage{eccv}

\usepackage{eccvabbrv}

\usepackage{graphicx}
\usepackage{booktabs}

\usepackage[accsupp]{axessibility}  

\usepackage{hyperref}

\usepackage{orcidlink}

\usepackage{algorithm}
\usepackage{algorithmic}
\usepackage[table]{xcolor}
\usepackage{multirow}
\usepackage{array}
\usepackage{nicematrix}
\usepackage{makecell}
\usepackage{wrapfig}

\definecolor{lightblue}{RGB}{230,245,255}
\definecolor{red}{RGB}{255,0,0}

\begin{document}

\title{AdaptToken: Entropy-based Adaptive Token Selection for MLLM Long Video Understanding} 

\titlerunning{AdaptToken}

\author{Haozhe Qi\inst{1,2}$^*$ \and
Kevin Qu\inst{3} \and
Mahdi Rad\inst{1} \and
Rui Wang\inst{1} \and
Alexander Mathis\inst{2} \and
Marc Pollefeys\inst{1,3}
}

\authorrunning{H.~Qi et al.}

\institute{$^1$Microsoft Spatial AI Lab \quad $^2$EPFL \quad $^3$ETH Zurich}

\maketitle
\let\thefootnote\relax\footnotetext{$^*$Work done during an internship at Microsoft}

\begin{abstract}


Long video understanding remains challenging for Multi-modal Large Language Models (MLLMs) due to high memory costs and context-length limits. Prior approaches mitigate this by scoring and selecting frames/tokens within short clips, but they lack a principled mechanism to (i) compare relevance across distant video clips and (ii) stop processing once sufficient evidence has been gathered. 
We propose AdaptToken, a training-free framework that turns an MLLM’s self-uncertainty into a global control signal for long-video token selection. AdaptToken splits a video into groups, extracts cross-modal attention to rank tokens within each group, and uses the model's response entropy to estimate each group's prompt relevance. This entropy signal enables a global token budget allocation across groups and further supports early stopping (AdaptToken-Lite), skipping the remaining groups when the model becomes sufficiently certain. Across four long-video benchmarks (VideoMME, LongVideoBench, LVBench, and MLVU) and multiple base MLLMs (7B--72B), AdaptToken consistently improves accuracy (e.g., +6.7 on average over Qwen2.5-VL 7B) and continues to benefit from extremely long inputs (up to 10K frames), while AdaptToken-Lite reduces inference time by about half with comparable performance. Project page: \url{https://haozheqi.github.io/adapt-token}

\end{abstract}

\section{Introduction}
\label{sec:intro}

Understanding long videos is essential for applications such as embodied AI assistants~\cite{bonnetto2025epfl} and multi-modal web agents~\cite{fan2024videoagent}. Recent multi-modal large language models (MLLMs)~\cite{bai2025qwen2,chen2024expanding,zhang2024video,zhang2025videollama} have achieved strong question-answering and instruction-following performance on short clips. However, they still struggle with long videos (e.g., hour-long inputs), largely because memory demands and context-length limits constrain both the resolution and the number of frames that can be processed~\cite{yao2026towards}.

To mitigate these limitations, recent works~\cite{liang2024keyvideollm,tang2025adaptive,hu2025m,yao2025generative} pre-select relevant or informative frames based on the text prompt and visual content before feeding them into MLLMs. However, frame-level selection often retains substantial irrelevant regions within the chosen frames, which can dilute the useful signal and hurt performance. Alternatively, token-level methods~\cite{cheng2025scaling,islam2025bimba,wang2025adaretake,zhang2025flexselect} directly select or compress the visual tokens provided to the MLLM, typically using cross-modal attention from vision-text encoders~\cite{radford2021learning} or MLLM layers to score token importance. While this fine-grained approach reduces redundancy compared to frame-level selection, two challenges remain. First, token importance is usually computed within individual frames or short clips, lacking a global criterion to allocate tokens across distant clips. Second, the selector typically still processes all sampled frames, incurring unnecessary computation when the MLLM could answer accurately using fewer groups.

\begin{wrapfigure}{r}{0.5\textwidth}
    \centering
    \vspace{-.75cm}
    \includegraphics[width=\linewidth]{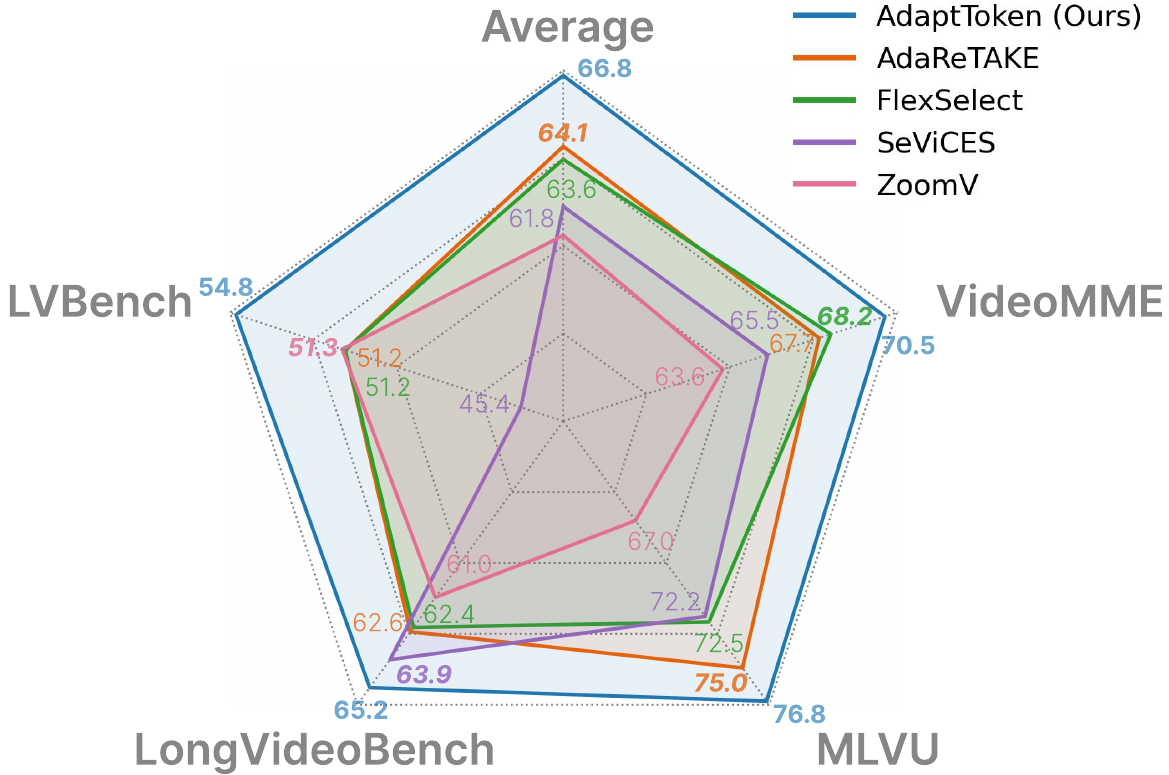}
    \caption{We propose AdaptToken, a flexible and efficient token selection strategy for long video understanding. Compared with state-of-the-art frame/token selection methods on several challenging long-video benchmarks, AdaptToken consistently delivers improved performance.}
    \vspace{-.7cm}
    \label{fig:net}
\end{wrapfigure}

\begin{figure}[t]
    \centering
    
    \includegraphics[width=\linewidth]{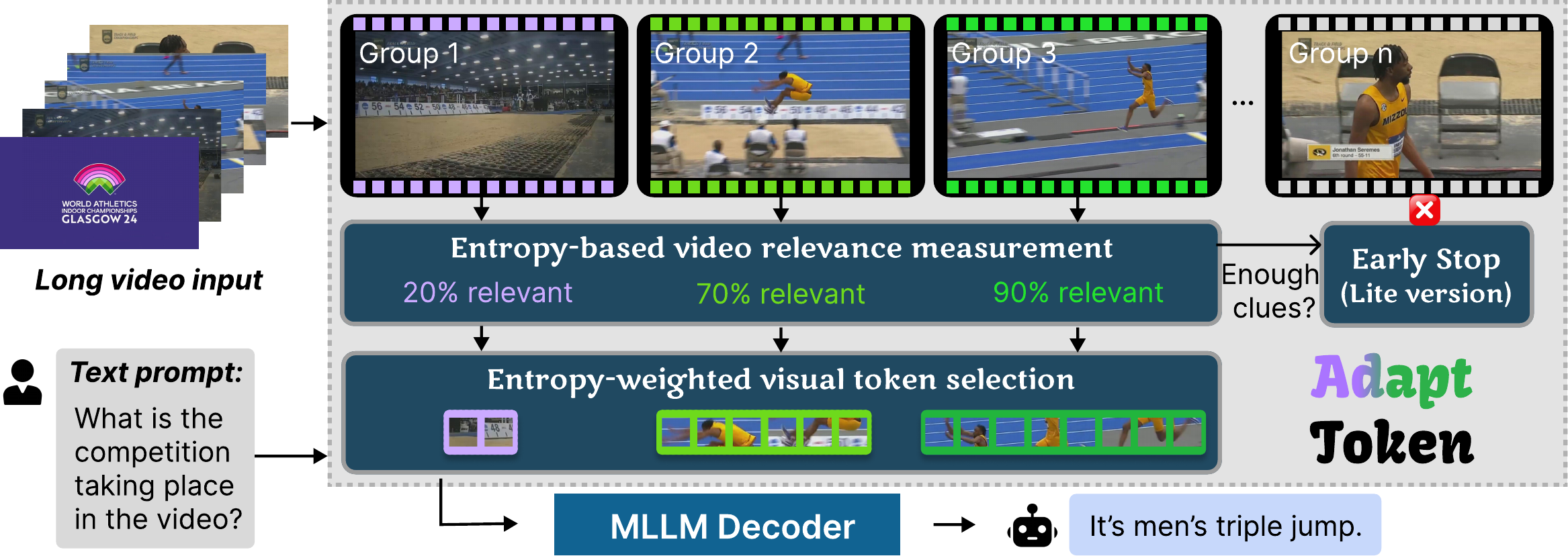}
    \caption{\textbf{Overall pipeline of AdaptToken.} AdaptToken processes long videos by dividing them into frame groups and selecting informative tokens within each group based on group relevance estimated from response entropy. It progressively gathers evidence across groups and stops processing once sufficient information has been collected.}
    \vspace{-.3cm}
    \label{fig:teaser}
\end{figure}

To address these limitations, we propose AdaptToken, a training-free long-video token selection framework suitable for various base MLLMs that (i) estimates token importance globally and (ii) decides when to stop examining additional video frames (Fig.~\ref{fig:teaser}). Specifically, AdaptToken processes the video group by group to avoid memory and context-length bottlenecks: within each group, it ranks visual tokens via cross-modal attention from a reference MLLM layer, yielding intra-group token relevance scores. To enable global comparisons across groups, we use the model's response entropy as a group-level uncertainty signal: groups that yield more confident (lower-entropy) responses are treated as more prompt-relevant and receive a larger share of the overall token budget.
The same entropy signal further supports early stopping (AdaptToken-Lite): once the model becomes sufficiently confident after examining a few groups, we skip the remaining groups to reduce computation.
Finally, token importance is not only attributed to prompt relevance. We improve token diversity and temporal coverage with a location-aware global token removal step that suppresses redundant tokens.

We integrate AdaptToken into multiple MLLM architectures~\cite{bai2025qwen2,chen2024expanding,Qwen3-VL} spanning model sizes from 7B to 72B parameters. Across four challenging long-video benchmarks, VideoMME~\cite{fu2025video}, LongVideoBench~\cite{wu2024longvideobench}, LVBench~\cite{wang2025lvbench}, and MLVU~\cite{zhou2025mlvu}, AdaptToken delivers consistent gains over the base MLLMs and prior frame/token selection methods (Fig.~\ref{fig:net}). Notably, thanks to global awareness during token selection and redundancy removal, AdaptToken scales to extremely long inputs and continues to improve performance even with 10K frames. To the best of our knowledge, prior MLLM long-video understanding methods~\cite{cheng2025scaling,li2024videochat} only report Needle-in-a-Haystack results rather than end-task accuracy at this scale.
Our early-stopping variant, AdaptToken-Lite, cuts average inference time by roughly half while maintaining comparable (or even better) performance. In summary, our contributions are threefold.

\begin{itemize}
 \item We propose AdaptToken, a training-free token selection framework that combines cross-modal attention with response entropy to assess token importance globally for long-video understanding.
 \item We further use response entropy to trigger early stopping, substantially accelerating inference while maintaining accuracy.
 \item We demonstrate AdaptToken's effectiveness across diverse MLLMs (multiple architectures and scales) and show consistent improvements on four long-video benchmarks.
\end{itemize}

\section{Related Works}

\subsection{Multi-modal Large Language Models}
Driven by the success of Large Language Models (LLMs)~\cite{achiam2023gpt,yang2025qwen3,cai2024internlm2}, multi-modal LLMs have emerged to extend text-only models with additional modalities~\cite{han2024onellm,girdhar2022omnivore,mckinzie2024mm1,comanici2025gemini}. Among them, video MLLMs~\cite{li2024llava, zhang2024video,wang2024internvideo2,li2024mvbench,xu2024slowfast} aim to provide robust and scalable solutions for understanding and reasoning over video data. A typical architecture comprises a visual encoder and a projection layer that maps visual features into the language embedding space, followed by an LLM backbone that processes the resulting multi-modal token sequence. However, the redundancy in video streams produces a large number of visual tokens, which quickly exhausts the context length and incurs substantial memory and compute overhead. AdaptToken mitigates this challenge by selecting a compact set of informative tokens.

\subsection{Long Video Understanding}
To process more frames, base MLLMs~\cite{zhang2024long,Qwen3-VL,wang2025internvideo2} typically extend the context window via long-video training or apply more aggressive temporal/spatial pooling. Nevertheless, efficiency remains a key bottleneck when handling large numbers of frames. Motivated by the redundancy in videos, recent work focuses on extracting prompt-relevant or informative content before feeding it into an MLLM. At the frame level, methods leverage CLIP~\cite{radford2021learning} vision-text encoders~\cite{liang2024keyvideollm,tang2025adaptive}, text-only LLMs~\cite{wang2025videotree}, or learned frame selectors~\cite{yu2024frame,yao2025generative} to identify relevant frames. However, selected frames may still contain substantial background or irrelevant regions, motivating feature/token-level compression based on cross-modal attention from vision-text encoders~\cite{wang2025seal,mohaiminul2025bimba} or MLLM layers~\cite{wang2025adaretake,zhang2025flexselect}. AdaptToken follows this line of work, aiming to improve efficiency while enabling globally informed token selection.

\subsection{Certainty Estimation for Model Responses}
\label{sec:certainty}
Humans are more likely to make incorrect statements when uncertain; analogously, recent work studies how to estimate an LLM's confidence to detect or mitigate hallucinations. Self-Evaluation~\cite{ren2023self} estimates confidence from the probability of yes/no tokens, while INTUITOR~\cite{zhao2025learning} learns dedicated confidence tokens via additional finetuning. Confidence can also be inferred directly from the output distribution using metrics such as token-level entropy and related uncertainty scores~\cite{fadeeva2024fact}, or self-certainty based on KL divergence from a uniform distribution~\cite{kang2025scalable}. These estimates have been used to improve test-time reasoning via multi-round voting~\cite{kang2025scalable,fu2025deep}. In contrast, AdaptToken leverages model certainty to assess group relevance and derive globally informed token importance for long-video token selection.

\section{Method}

AdaptToken consists of three main components. Given a long video, AdaptToken first splits it into multiple frame groups. For each group, we compute the group relevance (Section~\ref{sec:group_relevance}) and intra-group token importance to enable globally informed token selection (Section~\ref{sec:token_selection}). Next, we apply location-aware global token removal to the selected tokens to improve diversity and coverage beyond relevance (Section~\ref{sec:token_removal}). Finally, AdaptToken supports early stopping: once the model has gathered sufficient evidence, it stops examining additional groups to improve inference efficiency (Section~\ref{sec:early_stop}).


\subsection{Group Relevance with Response Entropy}
\label{sec:group_relevance}

Due to memory and context-length constraints, we cannot feed a large number of video frames (e.g., $>\!1$K frames) into an MLLM at once. A common practice is therefore to process either individual frames~\cite{hu2025cos} or small groups of frames~\cite{wang2025adaretake,zhang2025flexselect,pan2025timesearch} and select relevant frames/tokens within each group.
However, inter-group relevance is often poorly handled. Many methods rely on a binary decision: irrelevant frames/groups are either discarded entirely~\cite{hu2025cos} or aggressively compressed (e.g., into a single token~\cite{cheng2025scaling}).
As the number of groups grows, their contributions to a given prompt can vary substantially, making it beneficial to assess group-wise importance in a finer-grained and continuous manner.

\begin{wrapfigure}{r}{0.6\textwidth}
    \centering
    \vspace{-.85cm}
    \includegraphics[width=\linewidth]{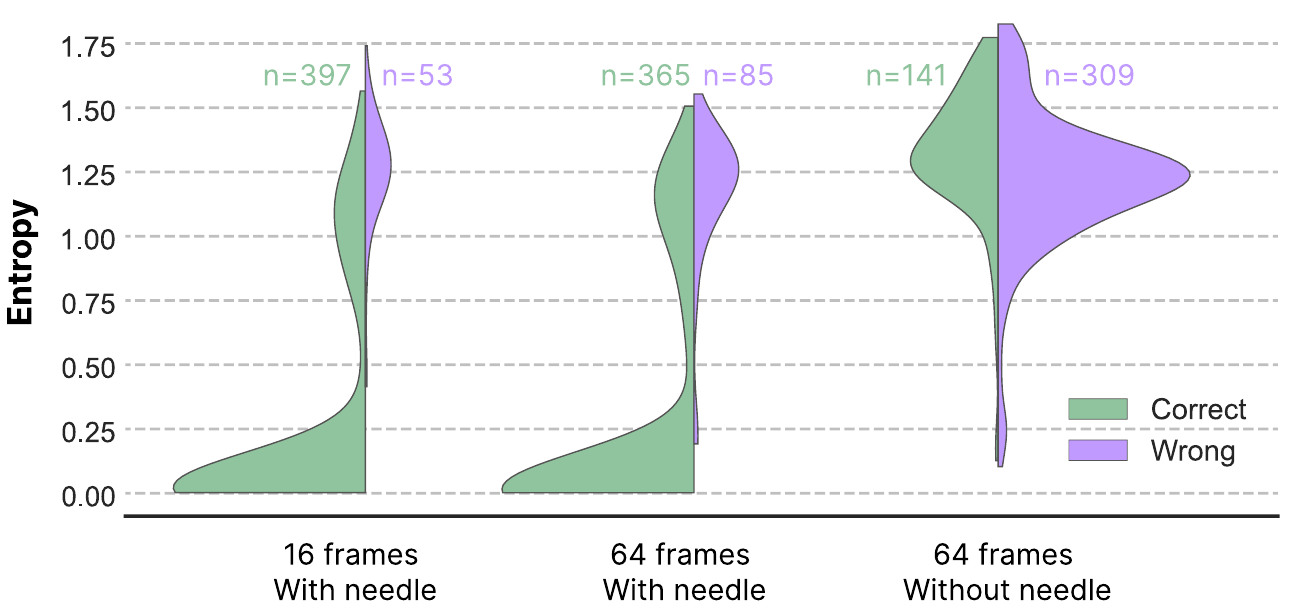}
    \caption{\textbf{Needle-in-a-Haystack experiments based on InternVL2.5 8B.} Response entropy distributions for correct vs. incorrect predictions under varying numbers of input frames, with and without needles.}
    \vspace{-.7cm}
    \label{fig:entropy}
\end{wrapfigure}

Inspired by recent progress in self-certainty estimation, we use an MLLM's response probabilities to quantify frame-group relevance. Prior work primarily uses self-certainty to improve test-time reasoning in text-only LLMs~\cite{kang2025scalable,fu2025deep}, e.g. by evaluating multiple reasoning traces and voting for the final answer. In contrast, we extend certainty estimation to MLLMs with video inputs and use it to assess the relevance of the input frames to the prompt.

Specifically, given a prefix of input tokens $\{x_1,\ldots,x_i\}$, the LLM backbone autoregressively produces a probability distribution $P_i \in \mathbf{R}^D$ over the next token, where $D=|\mathcal{D}|$ is the vocabulary size. The next token index is

\begin{equation}
    t_{i+1} = \operatorname*{argmax}_{j\in\{1,\ldots,D\}} P_i(j),
\end{equation}

and the corresponding token embedding is $x_{i+1}=Embeds[t_{i+1}]$, where $Embeds[\ ]$ denotes an embedding lookup table. Beyond selecting the next token, $P_i$ also reflects the model's belief about the output. Denoting the probability of vocabulary item $j$ as $P_i(j)$, the token entropy~\cite{kang2025scalable} is

\begin{equation}
\label{equ:entropy}
    e_i = - \sum_{j=1}^{D} P_i(j) \log P_i(j).
\end{equation}

Entropy measures the uncertainty of a probability distribution: higher entropy indicates greater uncertainty. We therefore define token certainty as $c_i=-e_i$, and follow Fu et al.~\cite{fu2025deep} by averaging the certainties of the lowest 10\% generated tokens to obtain the response-certainty score.

\begin{equation}
\label{equ:certainty}
    C = \frac{1}{|\mathcal{G}|}\sum_{i\in\mathcal{G}}c_i.
\end{equation}

Here, $\mathcal{G}$ denotes the set of generated tokens with the lowest 10\% certainty scores. We use $C$ to quantify the relevance of the input frames to the text prompt, based on the hypothesis that higher response certainty indicates that the model has observed more prompt-relevant evidence.

To validate this hypothesis, we first conduct Needle-in-a-Haystack experiments by injecting a needle frame into random videos and asking questions tied to that needle. As shown in Fig.~\ref{fig:entropy}, when there is no evidence (i.e., no needle frame) in the input, the MLLM remains highly uncertain even when it guesses correctly, suggesting that the model does not observe prompt-relevant evidence and is instead making a random guess. By contrast,  when the input contains the required evidence (i.e., the needle frame), the MLLM typically produces lower-entropy responses. 
A subset of cases still exhibits high entropy despite the presence of the needle, indicating that the model fails to retrieve or attend to the relevant frame and therefore behaves similarly to the no-evidence condition, making incorrect answers more likely.
Moreover, reducing the number of input frames from 64 to 16 decreases the fraction of high-entropy responses, consistent with easier retrieval under shorter contexts.

\begin{figure}[t]
    \centering
    \includegraphics[width=.85\linewidth]{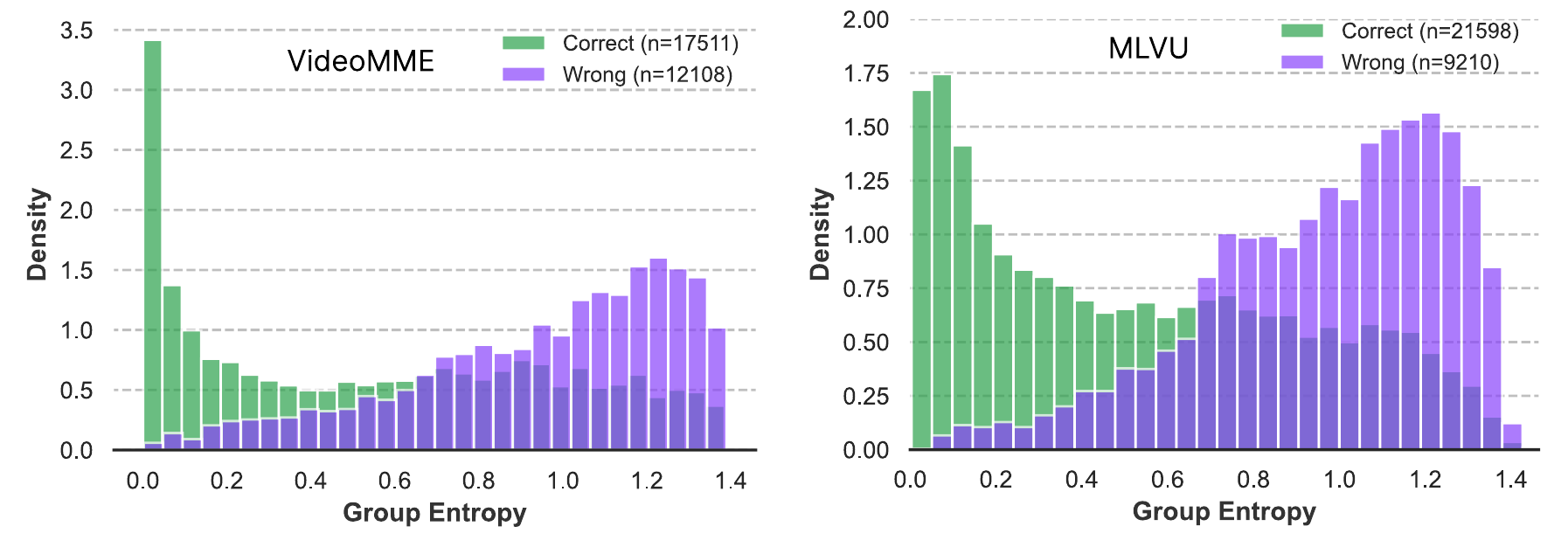}
    \vspace{-.15cm}
    \caption{\textbf{Real-data entropy experiments on based on InternVL2.5 8B.} Response-entropy distributions for correct vs. incorrect predictions on real-world benchmarks (VideoMME and MLVU).}
    \vspace{-.3cm}
    \label{fig:entropy_real}
\end{figure}

We further examine this relationship on real-world benchmarks (Fig.\ref{fig:entropy_real} and Appendix Fig.~\ref{fig:qwen}). Given a video with $N$ frames $\{f_1,\ldots,f_N\}$ sampled at a fixed FPS, we partition it into groups of at most $K$ frames, yielding $G=N//K+1$ groups $\{\mathcal{F}_1,\ldots,\mathcal{F}_G\}$. Within each group, frames are sampled with a stride $G$ so that each group spans the full video but with different temporal offsets. For example, group ${F}_g$ consists of:

\begin{equation}
\label{equ:frame_split}
    \mathcal{F}_g=\{f_g,f_{g+G},\ldots\}.
\end{equation}

We then process each group with the MLLM to obtain a response and its associated certainty score. Across benchmarks, higher certainty correlates with a higher probability of answering correctly, indicating that the corresponding frame group contains prompt-relevant evidence.

Compared to prior inter-group relevance measurements~\cite{hu2025cos,cheng2025scaling}, response certainty provides a quantitative and training-free signal for estimating inter-group relevance, which we leverage to derive globally informed token-importance scores.

\subsection{Entropy-guided Global Token Selection}
\label{sec:token_selection}

With the entropy-based response certainty measure, a video split into $G$ frame groups yields one certainty score $C_g$ per group $\mathcal{F}_g$, which captures group-level relevance. We then estimate token-level relevance within each group; importantly, token relevance and $C_g$ can be obtained in the same forward pass.

Token importance can be estimated either using pretrained vision-text encoders~\cite{wang2025internvideo2,mohaiminul2025bimba} or using attention from MLLM layers~\cite{wang2025adaretake,zhang2025flexselect,wang2024retake}. We adopt the latter because MLLMs better capture the semantics of complex text prompts. Concretely, for a given MLLM, we select a late layer with strong retrieval performance (identified via a layer-wise Needle-in-a-Haystack experiment~\cite{zhang2025flexselect}) and use it to compute cross-modal attention.
During the same forward pass used to compute group certainty, we extract the visual key embeddings $\{k_1,\ldots,k_V\}$ and text query embeddings $\{q_1,\ldots,q_T\}$ from the selected LLM layer to compute cross-modal attention. We then derive token relevance scores for the visual tokens, $R=\{r_1,\ldots,r_V\}$, by aggregating attention weights over heads and taking the maximum across text queries:

\begin{equation}
\label{equ:relevance}
    r_v = \max_{t\in\{1,\ldots,T\}} \sum_{h=1}^{H} \mathrm{Attn}_{h}(q_t, k_v), \quad v\in\{1,\ldots,V\}.
\end{equation}

\begin{figure}[t]
    \centering
    \includegraphics[width=\linewidth]{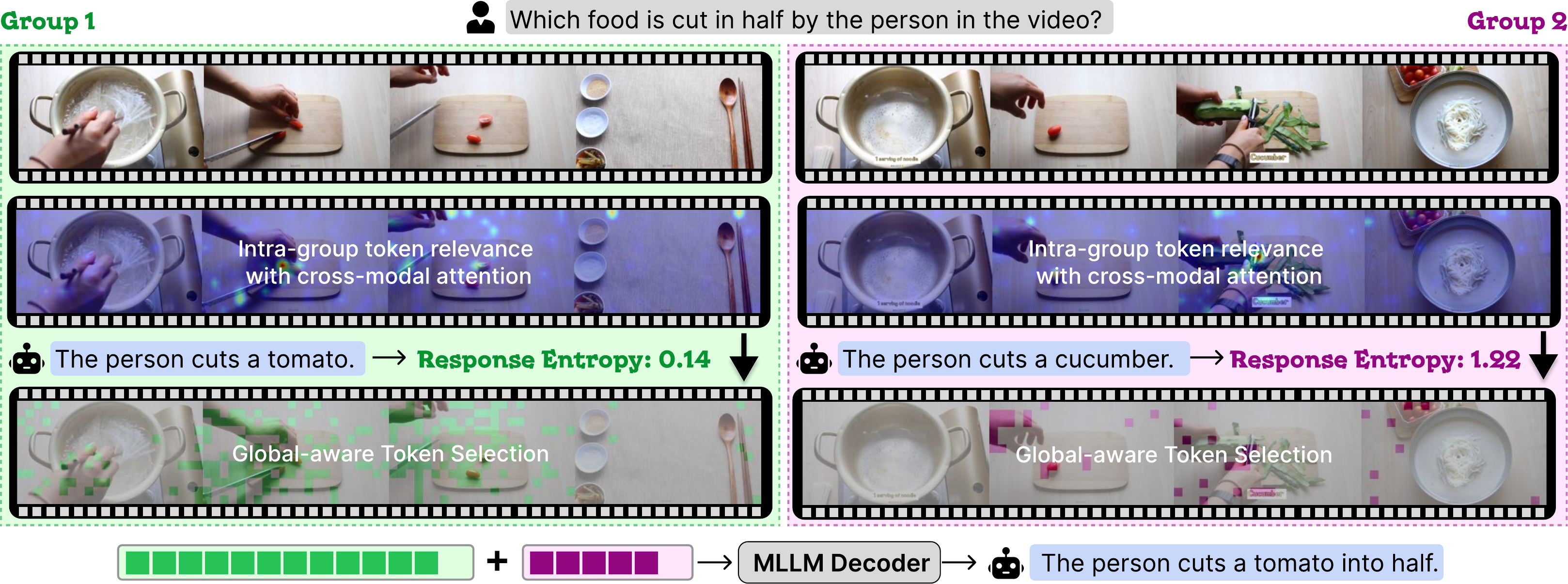}
    \caption{\textbf{Visualization of AdaptToken token selection.} Two frame groups are presented side by side. For each clip, we first estimate intra-group token relevance via cross-modal attention (heatmaps in the second row), and group-level relevance via response entropy, which measures the model’s answer confidence. Based on these signals, we perform global-aware token selection, adaptively allocating a larger token budget to groups that are more relevant to the text prompt (colored masks in the third row). The resulting token set is compact yet information-dense, improving both accuracy and inference efficiency for long-video understanding.}
    \vspace{-.3cm}
    \label{fig:detail}
\end{figure}

Here, $H$ denotes the number of attention heads and $\mathrm{Attn}_{h}(q_t, k_v)$ is the cross-modal attention weight from query $q_t$ to key $k_v$ in head $h$. We compute $R_g$ for each frame group to obtain intra-group token-importance scores. To enable global-aware token selection, we adopt a two-stage token-allocation strategy. We first set an overall token budget $B$ for the entire video, and then allocate a group-level budget $B_g$ for each frame group $\mathcal{F}_g$ based on its certainty score $C_g$:

\begin{equation}
\label{equ:budget}
    B_g = B\times \Bigl(\mathrm{Softmax}(\{C_1,\ldots,C_G\}/\tau)\Bigr)_g.
\end{equation}

Here, $\tau$ is a temperature parameter controlling the sharpness (fixed to $2$ across all experiments). Given $B_g$, we select the top-$B_g$ visual tokens in group $g$ according to $R_g$: more relevant groups (higher $C_g$) retain more informative tokens. Finally, since token location is important for prediction, we also keep the corresponding positional embeddings for the selected tokens.

\subsection{Location-aware Global Token Removal}
\label{sec:token_removal}

Our two-stage token allocation strategy selects prompt-relevant tokens globally. However, beyond relevance, effective long-video understanding also requires diversity and coverage~\cite{tang2025adaptive,cheng2025scaling}. For instance, neighboring groups $\mathcal{F}_g$ and $\mathcal{F}_{g+1}$ may contain highly similar content. Processing both groups can yield similar certainty scores $C$ and token relevance scores $R$, causing redundant tokens to be selected.

To mitigate this redundancy, we introduce a token-removal step. After selecting tokens $\{x_1,\ldots,x_B\}$ via two-stage allocation, we compute the pairwise cosine similarity of token features, $S^f \in \mathbf{R}^{B\times B}$. Since redundancy often arises from temporal proximity, we also record the global frame indices of the selected tokens and compute a temporal similarity term. We first normalize frame indices to $[0,1]$, yielding $\{d_1,\ldots,d_B\}$, and then define
$S^d_{i,j} = \exp\bigl( - (d_i - d_j)^2/\sigma\bigr)$,
where $\sigma$ controls how quickly similarity decays with temporal distance.
We combine the two similarities as

\begin{equation}
\label{equ:similarity}
    S = S^f + S^d.
\end{equation}

Token removal is performed iteratively by discarding tokens with the highest mutual similarity. In practice, we initially select a slightly larger set (10\%) than the target budget $B$ and then reduce it back to $B$ using this redundancy-removal procedure. We further ablate different removal ratios in Appendix Table~\ref{tab:removal}.

\begin{algorithm}[t]
  \caption{AdaptToken for Token Selection}
  \label{alg:adapttoken}
  \begin{algorithmic}
    \STATE \textbf{Inputs:} Video frames $\{f_1,f_2,...,f_N\}$, text prompt $\mathcal{T}$, overall token budget $B$
    \STATE Encode $\mathcal{T}$ into text tokens $\{x^t_1,...,x^t_T\}$ using vocabulary embeddings $Embeds$
    \STATE Split the video into frame groups $\{\mathcal{F}_1,...,\mathcal{F}_{G}\}$ using Eq.~\ref{equ:frame_split}
    \STATE Define the group input order with maximum margin: $\hat{G}=[1, G/2, G/4, 3G/4, ...]$
    \STATE Initialize the confidence counter $count = 0$
    \REPEAT
    \FOR{$g$ \textbf{in} $\hat{G}$}
        \STATE Encode $\mathcal{F}_g$ into visual tokens $\{x^v_1,...,x^v_V\}$ using the MLLM vision encoder
        \STATE Feed $\{x^t_1,...,x^t_T\}$ and $\{x^v_1,...,x^v_V\}$ into the MLLM decoder
        \STATE Compute the group-wise certainty $C_g$ using Eq.~\ref{equ:entropy} and Eq.~\ref{equ:certainty}
        \STATE Compute the token-wise prompt relevance $R_g$ using Eq.~\ref{equ:relevance}
        \IF{$C_g > C^*$ \textbf{and} early stopping is enabled (AdaptToken-Lite)}
            \STATE $count \leftarrow count + 1$
        \ENDIF
    \ENDFOR
    \UNTIL{$count \geq 3$}

    \STATE Update $\hat{G}$ to skip the rest groups. Compute the group token budget $B_g$ using Eq.~\ref{equ:budget}
    
    \FOR{$g$ \textbf{in} $\hat{G}$}
        \STATE Select the top-$B_g$ tokens according to $R_g$
    \ENDFOR
    
    \STATE Compute mutual token similarity matrix $S$ using Eq.~\ref{equ:similarity}
    \STATE Iteratively remove the top $0.1B$ most similar tokens based on $S$
    
    \STATE Aggregate the remaining tokens and feed them into the MLLM decoder
  \end{algorithmic}
\end{algorithm}

\subsection{Inference Efficiency with Early-stopping}
\label{sec:early_stop}

Splitting the whole video into frame groups largely reduces the memory requirement and also slightly improves the inference speed since the model attentions are computed with shorter lengths. However, all the frame groups still need to be processed and examined, which linearly increases the inference time as the number of frame groups increases. For a super-long video, the MLLM may already gather enough clues using some frame groups without going through the whole video. However, this has been barely addressed by existing frame/token selection methods~\cite{wang2025adaretake,zhang2025flexselect,cheng2025scaling}. 

\begin{table}[t]
\centering
\scriptsize
\caption{\textbf{Comprehensive evaluation across long-video benchmarks.} Results are grouped by model family. Best performance within each block is shown in bold. The first two blocks report representative closed-source and open-source MLLMs as reference points. Subsequent blocks list each base MLLM followed by the corresponding integrated methods.}
\vspace{-.1cm}
\begin{tabular}{lcccccc}
\toprule
\textbf{Model} & \textbf{LLM Size} & \textbf{VideoMME} & \textbf{MLVU} & \textbf{LongVideoBench} & \textbf{LVBench} \\
\midrule
GPT-4o~\cite{openai2024gpt4ocard} & - & 71.9 & 64.6 & 66.7 & 27.0 \\
Gemini-1.5-Pro~\cite{team2023gemini} & - & 73.2 & - & 64.0 & 65.7 \\
Gemini-2.5-Pro~\cite{comanici2025gemini} & - & 84.3 & - & - & 78.7 \\
\midrule
mPLUG-Owl3~\cite{ye2024mplug} & 7B & 59.3 & 63.7 & 52.1 & - \\
NVILA~\cite{liu2025nvila} & 8B & 64.2 & 70.1 & 57.7 & - \\
ByteVideoLLM~\cite{wang2025dynamic} & 14B & 64.6 & 70.1 & 70.1 & - \\
TPO~\cite{li2025temporal} & 7B & 65.6 & 71.1 & 60.1 & - \\
VideoLLaMA3~\cite{zhang2025videollama} & 7B & 66.2 & 73.0 & 59.8 & 45.3 \\
ViLAMP~\cite{cheng2025scaling} & 7B & 67.5 & 72.6 & 61.2 & 45.2 \\
\midrule
InternVL2.5~\cite{chen2024expanding} & 8B & 64.2 & 68.9 & 59.5 & 43.4 \\
\textit{w/} ZoomV~\cite{pan2025timesearch} & 8B & 64.4 & 70.0 & 63.3 & 51.5 \\
\textit{w/} Triumph~\cite{suo2025trial} & 8B & 65.4 & 70.0 & 60.7 & 46.6 \\
\textit{w/} SeViCES~\cite{sheng2025sevices} & 8B & 64.7 & 72.1 & 61.7 & 46.7 \\
\textit{w/} FlexSelect~\cite{zhang2025flexselect} & 8B & 67.0 & 71.9 & 60.1 & 49.7 \\
\rowcolor{lightblue}
\textit{w/} AdaptToken-Lite (Ours) & 8B & 68.1 & \textbf{74.4} & \textbf{63.8} & 51.3 \\
\rowcolor{lightblue}
\textit{w/} AdaptToken (Ours) & 8B & \textbf{68.3} & 74.1 & 63.7 & \textbf{52.1} \\
\midrule
Qwen2.5-VL~\cite{bai2025qwen2} & 7B & 65.4 & 70.2 & 59.5 & 45.3 \\
\textit{w/} TimeSearch-R~\cite{pan2025timesearchr} & 7B & 66.6 & 71.5 & 60.1 & - \\
\textit{w/} ZoomV~\cite{pan2025timesearch} & 7B & 63.6 & 67.0 & 61.0 & 51.3 \\
\textit{w/} AdaReTAKE~\cite{wang2025adaretake} & 7B & 67.7 & 75.0 & 62.6 & 51.2 \\
\textit{w/} SeViCES~\cite{sheng2025sevices} & 7B & 65.5 & 72.2 & 63.9 & 45.4 \\
\textit{w/} FlexSelect~\cite{zhang2025flexselect} & 7B & 68.2 & 72.5 & 62.4 & 51.2 \\
\rowcolor{lightblue}
\textit{w/} AdaptToken-Lite (Ours) & 7B & 69.8 & 76.3 & 65.1 & 53.3 \\
\rowcolor{lightblue}
\textit{w/} AdaptToken (Ours) & 7B & \textbf{70.5} & \textbf{76.8} & \textbf{65.2} & \textbf{54.8} \\
\midrule
Qwen2.5-VL~\cite{bai2025qwen2}             & 72B & 73.4 & 76.3 & 66.2 & 47.3 \\
\textit{w/} AdaReTAKE~\cite{wang2025adaretake} & 72B & 73.5 & 78.1 & 67.0 & 53.3 \\
\textit{w/} FlexSelect~\cite{zhang2025flexselect} & 72B & 74.4 & 76.6 & 66.4 & 56.6 \\
\rowcolor{lightblue}
\textit{w/} AdaptToken (Ours) & 72B & \textbf{76.1} & \textbf{79.8} & \textbf{70.5} & \textbf{59.7} \\
\midrule
Qwen3-VL~\cite{Qwen3-VL} & 8B & 71.4 & 78.1 & 65.9 & 58.0 \\
\rowcolor{lightblue}
\textit{w/} AdaptToken (Ours) & 8B & \textbf{73.8} & \textbf{79.3} & \textbf{66.9} & \textbf{60.6} \\
\bottomrule
\end{tabular}

\label{tab:main}
\end{table}

With group certainty $C_g$, we can identify when to stop examining new frame groups with no additional computations needed, which largely improves the inference efficiency while keeping comparable performance. Specifically, we show that a frame group is likely to contain many prompt-relevant clues (Fig.~\ref{fig:entropy}) when its certainty score is sufficiently high, which provides a signal for deciding whether the model has gathered enough evidence. Only one group with high certainty may carry partial clues. Therefore, we stop reviewing new groups once multiple frame groups have achieved high enough certainties. Based on the MLLM entropy analyses, we find that the range of response entropy remains stable across Needle-in-a-Haystack experiments (Fig.~\ref{fig:entropy}), real benchmarks (Fig.~\ref{fig:entropy_real}), and different MLLMs (Appendix Fig.~\ref{fig:qwen}). Accordingly, we use the stopping criterion that three frame groups with entropy below $C^* = 0.75$, which trades off computation and reliability by demanding confirmation under multiple groups and generalizes across benchmarks and MLLMs. We ablate other settings in Appendix Table~\ref{tab:early_stop}, demonstrating robustness to different hyperparameter choices.

We send frame groups to the MLLM in a maximum margin manner, i.e., $[\mathcal{F}_1,\mathcal{F}_{G/2},\mathcal{F}_{G/4},\mathcal{F}_{3G/4},...]$, to ensure the diversity of the frame groups. Once stopped, we simply use the already reviewed frame groups to perform our two-stage token allocation and token redundancy removal methods to select tokens.

The final selected tokens contain abundant text-prompt-relevant information with high diversity and coverage, which provides a globally consistent compact representation for the original video. We aggregate them following the temporal order and send them again to the MLLM to obtain the final response. We summarize the details of AdaptToken in Algorithm~\ref{alg:adapttoken}.

\vspace{-.1cm}
\section{Experiments}

\subsection{Benchmarks and Models}

We evaluate AdaptToken on four public long-video benchmarks commonly used by other frame/token selection methods: VideoMME~\cite{fu2025video}, MLVU~\cite{zhou2025mlvu}, LongVideo-Bench \cite{wu2024longvideobench}, and LVBench~\cite{wang2025lvbench}. VideoMME, MLVU, and LongVideoBench include both short and long videos, spanning durations from a few seconds to over an hour and covering diverse tasks (e.g., topic reasoning, anomaly recognition, and video summarization), which tests robustness across video types and lengths. LVBench instead targets extremely long videos, with an average duration of 4,101 seconds and many samples exceeding two hours, providing a stringent test of scalability to ultra-long inputs. We evaluate under the LMMS-Eval framework~\cite{zhang2024lmms} and report the official metrics for each benchmark.

We integrate AdaptToken into multiple MLLMs, including InternVL2.5 8B~\cite{chen2024expanding}, Qwen2.5-VL 7B/72B~\cite{bai2025qwen2}, and Qwen3-VL 8B~\cite{Qwen3-VL}. We primarily benchmark on InternVL2.5 8B and Qwen2.5-VL 7B, two common backbones in prior frame/token selection work, to enable fair comparisons. We additionally evaluate on Qwen2.5-VL 72B to demonstrate scalability to larger models. Finally, Qwen3-VL provides a strong long-context base MLLM via ultra-long-context training; we show that AdaptToken further improves its performance while reducing time and memory consumption. We refer to the full method as AdaptToken, and to the variant with early stopping as AdaptToken-Lite. Additional implementation details are provided in the Appendix.

\subsection{Main results} 

\textbf{Comparison with state-of-the-art methods.}
We integrate AdaptToken into multiple base MLLMs and evaluate on the four long-video benchmarks in Table~\ref{tab:main}. To enable fair comparisons with prior frame/token selection methods, we focus on two commonly used backbones: InternVL2.5 8B~\cite{chen2024expanding} and Qwen2.5-VL 7B~\cite{bai2025qwen2}. Across all benchmarks, AdaptToken yields larger improvements over the corresponding base MLLMs than competing approaches, demonstrating strong effectiveness for long-video understanding.

For Qwen2.5-VL 7B, prior methods attain their best performance on different benchmarks (e.g., FlexSelect on VideoMME, AdaReTAKE on MLVU, SeViCES on LongVideoBench, and ZoomV on LVBench). In contrast, AdaptToken improves performance consistently across all benchmarks, with an average gain of +6.7 over the Qwen2.5-VL 7B baseline, and surpasses the previous best results on each benchmark. The largest gains occur on LVBench and MLVU, which contain extremely long videos (often exceeding two hours), highlighting AdaptToken's robustness on ultra-long inputs. We observe a similar trend on InternVL2.5: AdaptToken outperforms all compared methods on each benchmark and improves the base model by +5.6 on average. Together, these results demonstrate that AdaptToken generalizes well across MLLM backbones.

\begin{wraptable}{r}{0.54\textwidth}
\centering
\scriptsize
\vspace{-1.cm}
\caption{\textbf{Accuracy and inference-time comparison between AdaptToken and AdaptToken-Lite.} Qwen2.5-VL 7B is used as the base MLLM. AdaptToken-Lite achieves comparable accuracy while reducing average inference time by approximately 50\%.}
\vspace{.1cm}
\begin{tabular}{lcccc}
\toprule
\multirow{2}{*}{Methods} & \multicolumn{2}{c}{\textbf{AdaptToken-Lite}} & \multicolumn{2}{c}{\textbf{AdaptToken}} \\
\cmidrule(lr){2-3} \cmidrule(lr){4-5}
 & Acc. & Time (s) & Acc. & Time (s) \\
\midrule
VideoMME & 69.8 & 8.6 & 70.5 & 17.8 \\
LongVideoBench & 65.1 & 11.0 & 65.2 & 18.2 \\
MLVU & 76.3 & 10.1 & 76.8 & 21.5 \\
LVBench & 53.3 & 19.3 & 54.8 & 32.8 \\
\bottomrule
\end{tabular}

\label{tab:early_stop_qwen}
\vspace{-.65cm}
\end{wraptable}

\textbf{Generalization to larger MLLMs.}
To assess scalability, we further integrate AdaptToken into Qwen2.5-VL 72B. As shown in Table~\ref{tab:main}, AdaptToken consistently improves performance across all benchmarks, achieving an average gain of +5.7 over the 72B baseline and outperforming all existing token-selection methods evaluated at this scale. This indicates that AdaptToken remains effective when applied to larger-scale MLLMs.

\textbf{Efficiency with early-stopping.} Our early-stopping strategy, described in Section~\ref{sec:early_stop}, enables the MLLM to skip uninformative frame groups once sufficient evidence has been collected, thereby reducing inference time. Table~\ref{tab:early_stop_qwen} and Appendix Table~\ref{tab:early_stop_internvl} report the average per-sample inference time for AdaptToken and AdaptToken-Lite.
Using Qwen2.5-VL 7B as the base MLLM, AdaptToken-Lite achieves accuracy comparable to AdaptToken (average difference: $-0.7$) while reducing inference time by approximately 50\% (e.g., from 17.8s to 8.6s on VideoMME). Notably, AdaptToken-Lite still outperforms the previous state of the art.
When applied to InternVL2.5 8B, AdaptToken-Lite matches or exceeds AdaptToken; in particular, it outperforms AdaptToken on MLVU and LongVideoBench, further validating the effectiveness of early stopping.


\textbf{Generalization to advanced MLLMs with long context length.} MLLMs such as Qwen2.5-VL~\cite{bai2025qwen2} and InternVL2.5~\cite{chen2024expanding} are trained and evaluated primarily on short video clips, with input limits of 25K tokens for Qwen2.5-VL and 64 frames (approximately 16K tokens) for InternVL2.5. By selecting a compact set of informative tokens, AdaptToken effectively increases relevant video content in the input, leading to improved performance. Most recent MLLMs, such as Qwen3-VL~\cite{Qwen3-VL}, are explicitly designed to handle long-context inputs. Qwen3-VL supports up to 2048 frames and 224K tokens, enabling strong baseline performance on long-video benchmarks (Table~\ref{tab:main}). We show that AdaptToken remains beneficial even in this setting. When applied to Qwen3-VL 8B and using up to 4096 input frames, AdaptToken yields an average improvement of +1.8 on the four benchmarks while substantially improving efficiency. On LVBench, AdaptToken requires 41 GB of GPU memory and 40.7s per sample on average, compared to 96 GB and 58.5s for the base Qwen3-VL 8B.

\subsection{Ablations}

\begin{wraptable}{r}{0.55\textwidth}
\centering
\scriptsize
\vspace{-1.15cm}
\caption{\textbf{Additive ablation study of AdaptToken components.} TS denotes token selection and TR denotes token removal. InternVL2.5 8B is used as the base MLLM. The final configuration (f) corresponds to the full AdaptToken.}
\vspace{.1cm}
\begin{tabular}{lccc}
\toprule
Method & Max frames & VideoMME & MLVU \\
\midrule
a. Base InternVL2.5 & 64 & 64.2 & 68.9 \\ 
b. Base InternVL2.5 & 256 & 64.3 & 67.2 \\
c. b+Group TS & 256 & 66.9 & 70.1 \\
d. c+Scale up frames & 1024 & 67.1 & 70.9 \\
e. d+Global TS & 1024 & 68.0 & 73.6 \\
f. e+Global TR & 1024 & 68.3 & 74.1 \\
\bottomrule
\end{tabular}

\label{tab:ablation}
\vspace{-.65cm}
\end{wraptable}

\textbf{Effectiveness of different components.} To identify the contributions of different components in AdaptToken, we conduct ablation studies presented in Table~\ref{tab:ablation}. We report results on VideoMME and MLVU, which cover a wide range of video durations. Starting from the InternVL2.5 baseline with 64 input frames (a), simply increasing the maximum number of frames to 256 (b) does not consistently improve performance and even degrades accuracy on MLVU (from 68.9 to 67.2). Adding group-wise token selection while constraining the final context length to a fixed budget $B$ and preserving prompt-relevant tokens (c) leads to a clear performance gain. With group-wise token selection in place, further scaling the maximum number of input frames to 1024 (d) yields modest improvements, despite incorporating substantially more video content. This suggests that different frame groups contribute unequally to the task and should not be treated uniformly. Incorporating entropy-guided global token selection (e) significantly outperforms group-wise selection alone by prioritizing more relevant groups. Finally, applying our global token removal procedure (f) further increases relevant token diversity and yields additional performance gains.

\textbf{Runtime breakdown analysis.}
AdaptToken-Lite substantially reduces inference time compared to AdaptToken while maintaining comparable performance (Table~\ref{tab:early_stop_qwen} and Appendix Table~\ref{tab:early_stop_internvl}). To identify the source of this speedup, we measure a runtime breakdown using Qwen2.5-VL 7B as the base MLLM. Specifically, we decompose AdaptToken into four stages. The group-wise inference stage, which computes response certainty and token relevance, dominates the overall runtime. Processing one frame group takes 1.05\,s on average, including 0.45\,s for visual feature encoding and 0.60\,s for LLM-backbone inference. Although the visual encoder is much smaller than the LLM, it incurs comparable latency due to full-attention computation. Consequently, the total cost of this stage scales approximately linearly with the number of groups.
The remaining three stages are comparatively inexpensive. Entropy-guided global token selection takes 0.07\,s given the group certainties and token-relevance scores from the first stage. Global token removal takes 0.55\,s. Finally, MLLM inference over the selected tokens takes only 0.31\,s due to the reduced input length.
Overall, these measurements indicate that group-wise inference is the primary computational bottleneck, especially as the number of frame groups increases. This explains why AdaptToken-Lite, which reduces the number of processed groups via early stopping, achieves large inference-time savings.

\begin{wraptable}{r}{0.6\textwidth}
\centering
\scriptsize
\vspace{-1.15cm}
\caption{\textbf{AdaptToken with more input frames.} Qwen2.5-VL 7B is used as the base MLLM. AdaptToken maintains strong performance as the number of input frames increases, demonstrating robustness to extremely long video inputs (up to 10K frames).}
\vspace{.1cm}
\begin{NiceTabular}{ccccc}
\toprule
Max frames & VideoMME & MLVU & LongVideoBench & LVBench  \\
\midrule
4096 & \textbf{70.5} & 76.8 & 65.2 & 54.8  \\ 
8192 & 70.2 & \textbf{77.1} & 65.5 & 54.9 \\
10000 & 70.1 & \textbf{77.1} & \textbf{65.8} & \textbf{55.6} \\ 
\bottomrule
\end{NiceTabular}

\label{tab:scaling}
\vspace{-.65cm}
\end{wraptable}

\textbf{Scaling to 10k frames.}
AdaptToken enables MLLMs to process extremely long videos under a fixed memory budget by selecting prompt-relevant tokens in a globally informed manner. To evaluate this capability, we increase the maximum number of input frames to 10K and report results in Table~\ref{tab:scaling}. To the best of our knowledge, existing frame- and token-selection methods~\cite{cheng2025scaling,li2024videochat} primarily report Needle-in-a-Haystack experiments at this scale, rather than end-task benchmark accuracy.
As the number of input frames increases, AdaptToken improves accuracy on three of the four benchmarks. Performance on VideoMME remains largely unchanged, which we attribute to the fact that the evidence required by its questions can typically be retrieved from fewer frames. In contrast, LVBench, with the longest average video length, benefits the most from the additional frames that our method is able to process.

\textbf{Comparison of certainty measures.} We adopt response entropy to estimate answer certainty, as it demonstrates strong capability in distinguishing relevant from irrelevant frame groups. As discussed in Section~\ref{sec:certainty}, alternative metrics are also applicable, such as response confidence~\cite{fu2025deep} and response KL divergence~\cite{kang2025scalable}. 

\begin{wraptable}{r}{0.57\textwidth}
\centering
\scriptsize
\vspace{-.3cm}
\caption{\textbf{Comparison of certainty measures.} AdaptToken achieves comparable performance when paired with different self-certainty scores. Qwen2.5-VL 7B is used as the base MLLM.}
\vspace{.1cm}
\begin{NiceTabular}{lcc}
\toprule
Method & VideoMME  & MLVU  \\
\midrule
Response KL-divergence & 70.0  & 76.5  \\ 
Response confidence & 70.1 & 76.7 \\ 
Response entropy & 70.5 & 76.8  \\
\bottomrule
\end{NiceTabular}

\label{tab:certainty}
\vspace{-.65cm}
\end{wraptable}

Table~\ref{tab:certainty} compares these measures on VideoMME and MLVU. The results show only marginal differences among the metrics, likely because the generated answers are short (in contrast to long-form mathematical reasoning), where certainty estimates tend to diverge more substantially. Overall, response entropy performs slightly better than the alternatives. We emphasize that our contribution does not lie in proposing a new certainty metric, but in leveraging existing certainty measures to guide input clue estimation for MLLMs.

\begin{wraptable}{r}{0.6\textwidth}
\centering
\scriptsize
\vspace{-1.15cm}
\caption{\textbf{Comparison to different voting methods.} AdaptToken aggregates tokens from different frame groups in a global-aware manner and obtains better performance over the voting methods (i.e., majority voting, borda voting and weighted voting).}
\vspace{.1cm}
\begin{NiceTabular}{lWc{38pt}Wc{28pt}Wc{55pt}}
\toprule
Method  & VideoMME & MLVU & LongVideoBench \\
\midrule
Base InternVL2.5 & 64.2 & 68.9 & 59.5 \\ 
Majority voting & 64.6 & 70.9 & 60.0 \\ 
Borda voting & 64.9 & 71.1 & 60.7 \\
Weighted voting & 65.3 & 71.6 & 61.6 \\
AdaptToken (Ours) & \textbf{68.3} & \textbf{74.1} & \textbf{63.7} \\
\bottomrule
\end{NiceTabular}

\label{tab:voting}
\vspace{-.65cm}
\end{wraptable}

\textbf{Comparison to voting methods.} Model self-certainty has primarily been used in prior work for test-time reasoning in LLMs, where multiple reasoning traces are aggregated using voting-based schemes to produce a final answer~\cite{kang2025scalable, fu2025deep}. In contrast, AdaptToken leverages self-certainty to identify prompt-relevant clues within each frame group and performs globally-informed token selection before the final inference pass. We compare AdaptToken with several voting-based aggregation strategies in Table~\ref{tab:voting}. We implement and test three representative voting methods. Majority voting selects the most frequent response across groups. Weighted voting assigns weights to responses based on the corresponding group certainty. Borda voting ranks groups by certainty and assigns scores according to $v(r)=(N-r+1)^p$, where $r$ denotes the group rank and $p$ is set to 0.9 following Kang et al.~\cite{kang2025scalable}. While these voting methods improve over the base InternVL2.5 model, their gains are substantially smaller than those achieved by AdaptToken. This comparison underscores the advantage of using self-certainty for globally informed token selection rather than for post-hoc response aggregation.

\section{Conclusion}

We present AdaptToken, a training-free and model-agnostic framework for long-video understanding with MLLMs. AdaptToken uses the model's response entropy as a global relevance signal to allocate token budgets across frame groups and select informative visual tokens via cross-modal attention. Token similarities are also considered to improve the diversity and temporal coverage of the selected tokens. AdaptToken-Lite further uses the same signal for early stopping. Across four long-video benchmarks and multiple base MLLMs (7B-72B), AdaptToken consistently improves accuracy while scaling to extremely long inputs (up to 10K frames). AdaptToken-Lite cuts average inference time by about half with comparable performance. Future work includes improving frame grouping and traversal strategies for better early stopping, and exploring more effective intra-group token relevance scoring methods beyond cross-attention.




%
%
\bibliographystyle{splncs04}
\bibliography{main}

\newpage
\appendix
\onecolumn

\section{Benchmark details}

\textbf{VideoMME.} VideoMME~\cite{fu2025video} is a benchmark for video understanding with diverse video types and durations. It contains 900 videos, with 2,700 manually annotated multiple-choice question-answer pairs across 30 subfields. The dataset is split into three subsets by duration: short ($<$2 minutes), medium (4-15 minutes), and long (30-60 minutes).

\textbf{MLVU.} MLVU~\cite{zhou2025mlvu} spans the widest range of video lengths, from 3 minutes to 2 hours. It includes nine tasks, such as topic reasoning, anomaly recognition, video summarization, and plot question answering.

\textbf{LongVideoBench.} LongVideoBench~\cite{wu2024longvideobench} targets long-context video understanding with videos up to one hour. It contains 3,763 videos and 6,678 annotated multiple-choice questions across 17 categories, emphasizing referring reasoning that requires retrieving and analyzing multi-modal details from specific temporal segments.

\textbf{LVBench.} LVBench~\cite{wang2025lvbench} focuses on long-video understanding, with an average video duration of 4,101 seconds (4$\times$ longer than VideoMME and 5$\times$ longer than MLVU). It includes 1,549 annotated multiple-choice question-answer pairs covering tasks such as event understanding, key information retrieval, temporal grounding, and reasoning.

\section{Implementation details}

Except for the GPU-memory test on Qwen3-VL (conducted on an H200 GPU), all evaluations are performed under the LMMS-Eval framework~\cite{zhang2024lmms} on 8 H100 GPUs. InternVL2.5 and Qwen2.5-VL are evaluated with a maximum of 16K and 32K input tokens, respectively, following their official settings. We extend both context limits by $16\times$, which increases the maximum number of input frames to 1024 for InternVL2.5 and 4096 for Qwen2.5-VL. For the token budget $B$, we follow FlexSelect~\cite{zhang2025flexselect} and use 32 frame-equivalent tokens for fair comparisons, corresponding to 7,010 tokens for Qwen2.5-VL and 8,256 tokens for InternVL2.5. For Qwen3-VL, we increase the budget to 128 frame-equivalent tokens (32,768 tokens). We set the temporal similarity decay parameter to 0.3 in all experiments.
The Needle-in-a-Haystack evaluation uses the V-NIAH data introduced in LongVA~\cite{zhang2024long}. We combine its needle samples with randomly sampled videos from VideoMME to obtain the correct/incorrect distributions.

\section{Additional experiments}

\begin{figure}[ht]
    \centering
    \includegraphics[width=.8\linewidth]{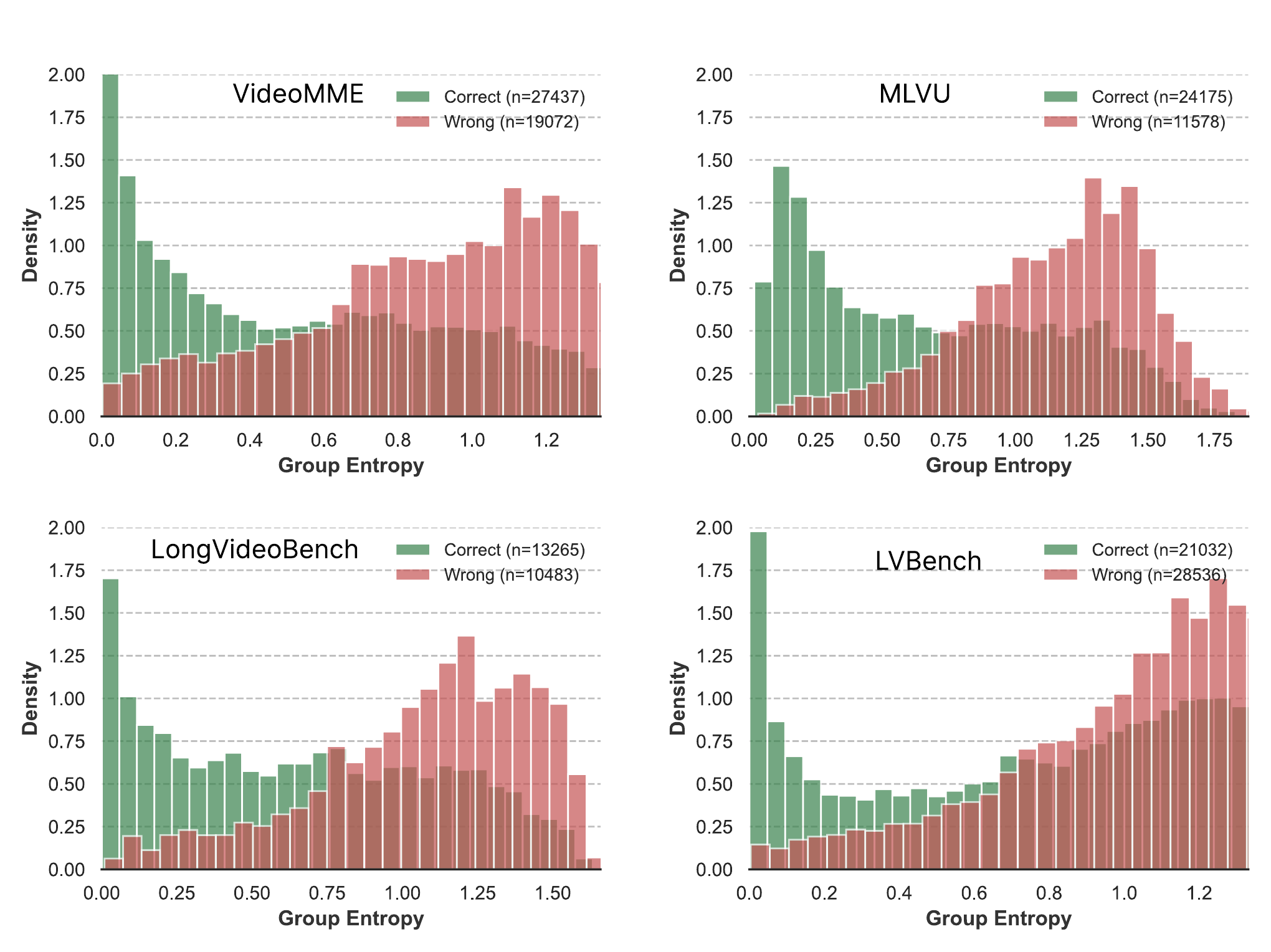}
    \caption{\textbf{Entropy distributions across benchmarks.} Qwen2.5-VL is used as the base MLLM.}
    \label{fig:qwen}
    \vspace{-.25cm}
\end{figure}

\begin{table}[h]
\centering
\small
\caption{\textbf{Accuracy and inference-time comparison between AdaptToken and AdaptToken-Lite.} InternVL2.5 8B is used as the base MLLM. AdaptToken-Lite achieves comparable or better accuracy while reducing inference time.}
\begin{tabular}{lcccccc}
\toprule
\multirow{2}{*}{Methods} & \multicolumn{3}{c}{AdaptToken-Lite} & \multicolumn{3}{c}{AdaptToken} \\
\cmidrule(lr){2-4} \cmidrule(lr){5-7}
 & Acc. & Groups & Time (s) & Acc. & Groups & Time (s) \\
\midrule
VideoMME       & 68.1 & 6.9  & 8.4 & 68.3  & 11.0 & 13.0 \\
LongVideoBench & 63.8 & 7.6  & 9.2  & 63.7 & 11.1 & 13.0 \\
MLVU           & 74.4 & 7.4  & 9.3  & 74.1 & 14.2 & 16.3 \\
LVBench        & 51.3 & 11.3 & 12.3 & 52.1 & 16.2 & 18.8 \\
\bottomrule
\end{tabular}

\label{tab:early_stop_internvl}
\end{table}


\textbf{Efficiency with early stopping for InternVL2.5.} In the main paper, we establish early-stopping effectiveness on Qwen2.5-VL 7B (Table~\ref{tab:early_stop_qwen}) using our early stopping version denoted as AdaptToken-Lite. Here, we keep the protocol unchanged and switch only the backbone to InternVL2.5 8B to test whether the gain is model-agnostic rather than architecture-specific. Specifically, we evaluate AdaptToken and AdaptToken-Lite under identical token budgets and benchmark splits, and report three metrics in Table~\ref{tab:early_stop_internvl}: task accuracy, the average number of inferred frame groups per sample, and end-to-end inference time. This appendix experiment therefore serves as a direct cross-backbone validation of the main-paper finding: the entropy-based stopping rule preserves accuracy while reducing computation. Consistent with the Qwen2.5-VL results, AdaptToken-Lite matches or slightly exceeds AdaptToken on LongVideoBench and MLVU, while requiring only about 65\% of the inference time. The reduction in processed groups correlates closely with inference time, confirming that group-wise MLLM inference is the dominant runtime cost.




\textbf{Group-wise entropy experiments with Qwen2.5-VL.} In the main paper (Fig.~\ref{fig:entropy_real}), we analyze group-wise response entropy on InternVL2.5 and show that lower entropy is associated with correct predictions, which motivates our entropy-based group ranking and early-stopping strategy. In this appendix experiment, we keep the same analysis protocol and change only the backbone to Qwen2.5-VL to test whether this entropy-correctness relationship is backbone-independent. Concretely, we construct 64-frame groups, compute group-wise response entropy under the same prompting/inference setup, and evaluate across all four benchmarks (VideoMME, LongVideoBench, MLVU, and LVBench). As shown in Fig.~\ref{fig:qwen}, correct and incorrect cases remain clearly separable in entropy space, and the distributional trend closely matches the InternVL2.5 results. This serves as a direct cross-backbone validation of the main-paper finding: response entropy is a stable uncertainty signal for long-video group selection, enabling the same early-stopping hyperparameters to transfer across models with little or no retuning.

\begin{table}[h]
\centering
\small
\caption{\textbf{Ablation on the token-removal rate.} AdaptToken is robust to different removal rates.}
\begin{NiceTabular}{lc}
\toprule
Method  & MLVU  \\
\midrule
Removing 0.05$B$ & 76.1  \\ 
Removing 0.1$B$ & 76.8 \\ 
Removing 0.2$B$ & 76.9  \\
Removing 0.3$B$ & 76.7  \\
\bottomrule
\end{NiceTabular}

\label{tab:removal}
\end{table}


\textbf{Ablation on the token-removal rate.} In the main paper, AdaptToken uses a fixed removal rate of 0.1 in all experiments to balance relevance and diversity after initial token selection. This appendix experiment isolates that design choice and asks whether performance depends critically on this specific setting. Following the same setup as the main results (same backbone, token budget $B$, and inference pipeline), we vary only the post-selection removal ratio and evaluate rates from 0.05$B$ to 0.3$B$ (Table~\ref{tab:removal}). The resulting accuracy differences are small, indicating that AdaptToken is not sensitive to the exact removal ratio within a broad range. This supports the main-paper configuration: 0.1 is a stable default that achieves near-best performance while avoiding extra hyperparameter tuning.

\begin{table}[h]
\centering
\small
\caption{\textbf{Ablation on early-stopping hyperparameters.}}
\begin{NiceTabular}{lcccccccc}
\toprule
Entropy threshold & 0.6  & 0.7  & 0.7  & 0.75 & 0.75 & 0.75 & 0.8  & 0.8 \\
Group threshold   & 1    & 1    & 2    & 1    & 2    & 3    & 3    & 4   \\
\midrule
MLVU              & 76.2 & 75.9 & 76.2 & 75.9 & 76.0 & 76.3 & 76.0 & 76.3 \\
\bottomrule
\end{NiceTabular}

\label{tab:early_stop}
\end{table}


\textbf{Ablation on early-stopping hyperparameters.} In the main paper, Adapt-Token-Lite uses a fixed early-stopping rule with an entropy threshold of 0.75 and a group threshold of 3 in all the experiments, showing a strong accuracy--efficiency trade-off. This appendix study examines whether that choice is robust. Using the same setup as the main experiments (same backbone, token budget, and inference pipeline), we vary only these two stopping hyperparameters and report MLVU accuracy in Table~\ref{tab:early_stop}. The entropy threshold controls how confidently a group must be answered to count as evidence, while the group threshold controls how many such confident groups are required before stopping. Lowering the group threshold makes stopping more aggressive, whereas increasing it makes stopping more conservative; similarly, a higher entropy threshold is permissive and a lower one is stricter. Results remain stable across a broad range, with the best performance achieved by multiple nearby settings (e.g., 0.75/3 and 0.8/4), indicating that the method is not sensitive to exact tuning. This directly supports the main-paper configuration: the default setting (0.75, 3) lies in a robust regime that transfers well without dataset-specific or MLLM-specific retuning.



\textbf{Ablation on frame-group construction.} In the main paper, AdaptToken relies on group-wise processing to scale long-video understanding, where each group serves as a unit for entropy estimation, token allocation, and early stopping. This appendix ablation isolates the group-construction strategy to test the effectiveness of our specific grouping design. Keeping the rest of the pipeline unchanged, we compare three input organizations on MLVU: (i) video-chunk inputs, which use contiguous local segments, (ii) continuous group inputs, which preserve temporal continuity with sequential grouping, and (iii) our marginal group inputs, which sample frames across the full video and process groups in maximum-margin order. The results (73.5, 74.0, and 74.4, respectively) show that our design consistently performs best.

\end{document}